\documentclass[letterpaper, 10 pt, conference]{ieeeconf}  

\IEEEoverridecommandlockouts      

\usepackage{amsmath} 
\usepackage{graphicx}
\usepackage{booktabs}
\usepackage{amssymb}
\usepackage[english]{babel}
\usepackage[noadjust]{cite}
\usepackage{units}
\usepackage{epstopdf}
\usepackage[font=small]{caption}
\usepackage{subcaption}
\usepackage{comment}
\usepackage{amsfonts}
\usepackage{bm}
\usepackage{algorithm}

\usepackage{amsthm}
\usepackage{algorithmic}

\usepackage{color}

\usepackage{xcolor}
\newcommand{\remove}[1]{}

\theoremstyle{plain}

\newtheorem{theorem}{Theorem}

\newtheorem{definition}[theorem]{Definition}

\newtheorem{remark}{Remark}

\title{\LARGE \bf
  Risk-aware Safe Control for Decentralized Multi-agent \\Systems via Dynamic Responsibility Allocation
}

\author{Yiwei Lyu$^1$, Wenhao Luo$^2$ and John M. Dolan$^3$
\thanks{$^*$This work was supported by the CMU Argo AI Center for Autonomous
Vehicle Research and the Faculty Research Grant award at UNC Charlotte.}
\thanks{$^1$The author is with the Department of Electrical and Computer Engineering, Carnegie Mellon University, Pittsburgh PA 15213, USA. Email: {\tt \small yiweilyu@andrew.cmu.edu}.}
\thanks{$^{2}$The author is with the Department of Computer Science, University of North Carolina at Charlotte, Charlotte NC 28223, USA. Email: {\tt \small wenhao.luo@uncc.edu}.}
\thanks{$^{3}$The author is with the Robotics Institute, Carnegie Mellon University, Pittsburgh PA 15213, USA. Email: {\tt \small jmd@cs.cmu.edu}}%
}

\begin{document}

\maketitle
\thispagestyle{empty}
\pagestyle{empty}

\begin{abstract}
Decentralized control schemes are increasingly favored  in various domains that involve multi-agent systems due to the need for computational efficiency as well as general applicability to large-scale systems. However, in the absence of an explicit global coordinator, it is hard for distributed agents to determine how to efficiently interact with others. In this paper, we present a risk-aware decentralized control framework that provides guidance on how much relative responsibility share (a percentage) an individual agent should take to avoid collisions with others while moving efficiently without direct communications. We propose a novel Control Barrier Function (CBF)-inspired risk measurement to characterize the aggregate risk agents face from potential collisions under motion uncertainty. We use this measurement to allocate responsibility shares among agents dynamically and develop risk-aware decentralized safe controllers. In this way, we are able to leverage the flexibility of robots with lower risk to improve the motion flexibility for those with higher risk, thus achieving improved collective safety.
We demonstrate the validity and efficiency of our proposed approach through two examples: ramp merging in autonomous driving and a multi-agent position-swapping game.

\end{abstract}

\section{Introduction}

In multi-agent systems where agents interact with each other in close proximity, the risk each individual agent faces, e.g. reflecting how likely one will collide with other agents, can differ based on various factors, and this information should also be reflected in controller design. In this work, we focus on both the direct and aggregated risks caused by existence of surrounding agents and agents' motion under uncertainty. Conventional risk evaluation methods for collision avoidance only consider agent positions and estimate the risk based solely on the inter-agent proximity~\cite{soriano2013multi}. However, it is important to also consider agent motion when evaluating risk~\cite{pierson2018navigating,toytziaridis2019data}, as two agents close to the same position moving away from or towards each other will have different collision likelihoods. 
Recently, 
Control Barrier Function (CBF) \cite{ames2019control} as a model-based approach has been widely studied to render a set forward invariant. CBF is often used to characterize safety in terms of collision avoidance between pairwise agents based on factors such as agent positions, motion, safety radius, and agent behavior conservativeness \cite{wang2017safety, luo2020multi}. It provides a means to constrain the robot motion, so that if the robot is initially inside the defined safe set, CBF can always ensure the robot stays within the safe set with formally provable guarantees. Existing works mostly use CBF as a constraint for optimization-based controllers, and nominal control is only revised when the system is approaching the boundary of the safe set~\cite{wang2016safety,wang2017safety}. Different from simply using CBF as a binary verification of whether the system is safe given the nominal control, in this work we propose a CBF-inspired risk evaluation framework to characterize to what extent the system is safe or unsafe, in order to make the best use of the information CBF provides.

\begin{figure}
    \centering
    \includegraphics[width = 1\linewidth]{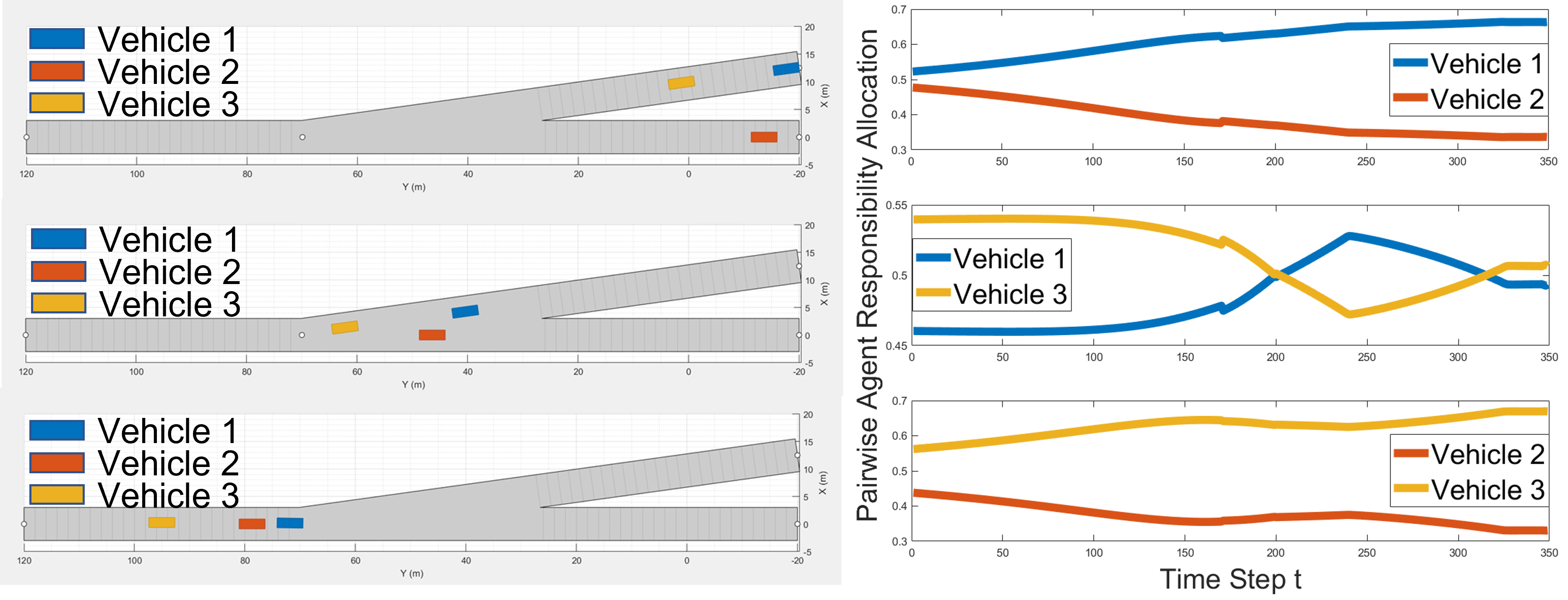}
    \caption{\footnotesize
    \label{responsibility-share}
   Dynamic responsibility allocation for pairwise agents in a ramp merging scenario, considering the risk from the potential collision under agent motion uncertainty in multi-agent interaction. The higher risk to which the agent is exposed, the less responsibility share is allocated to it, meaning the agent motion is more tightly constrained and forced to proceed with caution, compared to the agent facing lower risk. Detailed interpretations can be found in Sec. \ref{experiment}.B.}
   \vspace{-0.5cm}
\end{figure}

To keep a multi-agent system safe, the collision-free configuration is required for every pairwise agent during interactions. There has been some work addressing risk-aware control in dynamic environments with moving obstacles \cite{hakobyan2019risk,wang2020game}, by posing multiple pairwise safety constraints between the individual agent under control and its neighboring agents. However, we argue that to better characterize an agent's collision risk,  considering only the effect of its neighboring agents is not enough. We also need to consider the neighbors of the neighbors. The rationale behind this is that even if no collision occurs, the amount of risk the agent faces can vary depending on the existence and behavior of other agents in the shared environment. 
Therefore, another goal of this work is to propose a risk measurement framework that can accumulate the risk each agent faces to take neighbors of neighbors into account, providing situational awareness of the interactions among surrounding agents. The work in \cite{pierson2018navigating} constructs a risk level set using a hand-crafted cost function to quantify the influence of other agents positions and movement. However, it focuses on planning a safe trajectory for a single robot within the risk level set corresponding to a fixed risk threshold, which may be difficult to calibrate beforehand in order to keep a sizable admissible safe space in the interaction-intensive environment without 
exceeding the individual robot's risk tolerance.

On the other hand, realistic factors such as agents' motion uncertainty could also contribute to the risk imposed on individual agents.
Chance constraint in the form of $\text{Pr}(\cdot)\geq \alpha$ has been a popular tool to account for uncertainty by translating probabilistic
constraints into deterministic ones~\cite{huang2018hybrid,lyu2021probabilistic, luo2020multi}. 
Given a user-defined confidence level $\alpha\in (0,1)$, the chance constraint is able to justify whether or not the deterministic condition is met with the satisfying probability. However, it is unable to quantify accumulated risk on an individual agent from different surrounding agents it may face.
Compared to that, Conditional Value at Risk (CVaR) is considered as a more suitable tool for measuring risk from uncertainty~\cite{zhou2022risk}. 
With the user-defined confidence level $\alpha$, $\mbox{CVaR}_\alpha(X):= \min_{z\in \mathbb{R}} \mathbb{E}\left[  z+\frac{(X-z)^{+}}{1-\alpha}\right]$ \cite{rockafellar2002conditional} quantifies how bad the expected loss could be if the condition is violated. It maps the risk of uncertainty to a real number, allowing for better embedding of uncertainty information in risk measurement. Works in \cite{ahmadi2020risk,singletary2022safe} introduce the CVaR barrier function and Risk Control Barrier Function to enforce CVaR-safety and guarantee finite-time reachability to a desired set under uncertainty respectively. Different from these works, we focus on how to design risk-aware decentralized controllers by quantifying the cumulative risk the agents face in a crowded dynamic environment under uncertainty. 

With that, the motivation for developing decentralized safe controllers is twofold: centralized control of a large-scale multi-agent system is computationally expensive, and inter-agent communication is not always available. Different approaches have been explored to translate centralized safe control into a decentralized setting, by splitting the safety constraints and separately solving individual optimization problems with split constraints.
In this way, agents only need to make decisions based on local information, without the need to predict what others are going to do. 
For example, Voronoi Cells-based methods  are a common approach to achieve decentralized safe control by partitioning the joint state space of robots~\cite{zhu2022decentralized, pierson2020weighted}. However, since the constraint is posed over robots' states instead of control inputs being directly optimized, the resulting optimization problems can not be solved simply using off-the-shelf optimizers. 
On the other hand, CBF-based decentralized controllers 
are able to
translate constraints from the state space into control space, making the problem more directly solvable. Various criteria for the control space partition between agents have been employed, such as
agents' relative actuation limits~\cite{wang2016safety} and social personalities, e.g. aggressive vs. conservative, but it requires information associated with the criteria fixed and known beforehand.

 In this paper, we aim to develop a decentralized safe controller of individual agents that can adapt to the dynamic changes in the surrounding environments with other agents to achieve implicit coordination and improved collective safety. 
 In particular, our proposed algorithm dynamically allocates responsibility shares among agents, indicating the portion of constraint each agent is expected to respect compared to its pairwise companion, based on the accumulated risk it receives. The responsibility shares are further embedded into our CBF-based decentralized controller design to provide agents with situational awareness of the environment they are in.
 For each set of pairwise agents, a larger responsibility share is allocated to the agent facing less risk, and a smaller responsibility share is allocated to the agent with higher risk. The idea is to enforce tighter constraints on the motion of agents with higher risk to encourage them to proceed with more caution. Note that the goal here is to enable agents to make risk-aware decisions independently, rather than minimizing the entire system's risk that can often make agents over-conservative~\cite{scukins2021using,liu2022decentralized}.

Our \textbf{main contributions} are: \textbf{1)} By combining the concept of Conditional Value at Risk and Control Barrier Function-based safe control, we present a novel risk evaluation framework to quantify the cumulative risk the agents face in a crowded dynamic environment under uncertainty, naturally factoring in the influence of the neighbors of neighbors;
\textbf{2)} We demonstrate the use of the proposed CBF-inspired Risk Map (CBF-RM) to visualize and understand the risk evaluation in a dynamic environment; \textbf{3)} We 
formulate the decentralized CBF-based safe controller composition as a dynamic agent responsibility allocation problem associated with risk, resulting from an interpretable measure of relative safety and adaptive conservativeness among agents. Rigorous proof of safety guarantees is provided.

\section{Preliminaries of Control Barrier Function}

{Control Barrier Functions} (CBF) \cite{ames2019control} are used to define an admissible control space for safety assurance of dynamical systems.
One of its important properties is its forward-invariance guarantee of a desired safety set.
Consider a nonlinear system in control affine form: $\dot x = f(x)+g(x)u$,
where $x\in \mathcal{X}\subset \mathbb{R}^n$ and $u\in\mathcal{U}\subset \mathbb{R}^m$ are the system state and control input with $f$ and $g$ assumed to be locally Lipschitz continuous.
A desired safety set $\mathcal{H}$ can be denoted by a safety function $h(x)$: $\mathcal{H} =\{x \in \mathbb{R}^n : h(x)\geq 0\}$. Thus the control barrier function for the system to remain in the safety set can be defined as follows \cite{ames2019control}:
\begin{definition}
(Control Barrier Function) Given aforementioned dynamical system and the set $\mathcal{H}$ with a continuously differentiable function $h:\mathbb{R}^n\rightarrow \mathbb{R}$, then $h$ is a control barrier function (CBF) if there exists a class $\mathcal{K}$ function for all $x\in \mathcal{X}$ such that 
\begin{equation}\label{eq:cbf_def}
    \sup_{u\in\mathcal{U}} \ \{\dot{h}(x,u)\}\geq -\kappa \big(h(x)\big)
\end{equation}
\end{definition}
\noindent
 We selected the same class $\mathcal{K}$ function $\kappa (h(x))=\gamma h(x)$ as in~\cite{zeng2021safety,he2021rulebased}, where $\gamma\in\mathbb{R}^{\geq 0}$ is a CBF design parameter controlling system behaviors near the boundary of $h(x)=0$. Hence, the admissible control space in (\ref{eq:cbf_def}) can be redefined as $\mathcal{B}(x)=\{u\in\mathcal{U}:\dot{h}(x,u) + \gamma h(x)\geq 0\; \}$.
It is proved in \cite{ames2019control} that any controller $u\in\mathcal{B}(x)$ will render the safe state set $\mathcal{H}$ forward-invariant, i.e., if the system starts inside the set $\mathcal{H}$ with $x(t=0)\in \mathcal{H}$, then it implies $x(t)\in\mathcal{H}$ for all $t>0$ under controller $u\in\mathcal{B}(x)$.

\section{Method}

\subsection{Control Barrier Function-inspired Risk Evaluation}

\textbf{Research Question 1}: How can we quantify the accumulated risk an agent faces in multi-agent interactions under motion uncertainty? 

Consider a multi-agent system with a total number of agents $N \in \mathcal{N}$, in which every agent has access to observations of all agents' current positions and velocities, but no direct communication is available among agents. As in~\cite{lyu2021probabilistic,lyu2022adaptive, lyu2022responsibility}, we consider the particular choice of pairwise safety function $ h_{ij}(x)$, safety set $\mathcal{H}_{ij}(x)$, and admissible control space $\mathcal{B}_{ij}(x)$ for each agent pair as follows.
\begin{equation}\label{safety-function}
\begin{split}
        &\mathcal{H}_{ij}(x)=\{x\in\mathcal{X}: \;h_{ij}(x) = ||x_i-x_j||^2-R_{safe}^2\geq 0,\forall i\neq j\} \\
        &\mathcal{B}_{ij}(x)=\{u\in\mathcal{U}:\dot{h}_{ij}(x,u)\geq -\gamma (h_{ij}(x))\}
\end{split}
\end{equation} where $x_i,x_j \in \mathbb{R}^2$ for $i,j \in \{1,...,N\}$ are the positions of any pairwise agents $i$ and $j$. $ u = \{u_i,u_j\} \in \mathbb{R}^2$ is the joint control input of this particular agent pair, and $R_{safe}$ is the pre-defined safety margin.

Next, to quantify the risk between each pairwise agents from potential collision and motion uncertainty, we draw inspirations from CBF and propose the following pairwise safety loss function $L_{ij}(x,{u})$:
\begin{equation}
\begin{split}
      &L_{ij}(x,{u}) = -\mbox{CVaR}_{\alpha}(\dot{h}_{ij}(x,{u}))-\gamma h_{ij}(x)+c \\
       &= -\mbox{CVaR}_{\alpha}(2(x_i-x_j)^T({u}_i+\epsilon_i-{u}_j-\epsilon_j)) \\
       &-\gamma (||x_i-x_j||^2-R_{safe}^2)+c \\
        &= -2 (x_i-x_j)^T({u}_i-{u}_j) - 2 \cdot \mbox{CVaR}_{\alpha}((x_i-x_j)^T(\epsilon_i-\epsilon_j))\\
       &-\gamma (||x_i-x_j||^2-R_{safe}^2)+c  \\ &\mbox{(by CVaR Translational Invariance Property \cite{singletary2022safe})}
\end{split}
\end{equation}
where $c$ as a constant offset is a large number to ensure $L_{ij}(x,{u})$ is always positive to prevent unintended cancelling-out when being accumulated later. ${u}_i,{u}_j \in \mathbb{R}^2$ are the agent's current velocities. 
$\epsilon_i, \epsilon_j \sim \mathcal{N}(\hat{\epsilon}, \Sigma)$ are random Gaussian variables with known mean $\hat{\epsilon}\in\mathbb{R}^2$ and variance $\Sigma\in\mathbb{R}^{2\times2}$, representing the uncertainty in each vehicle's motion. $\gamma$ is the CBF design factor representing how aggressive the pairwise agents are \cite{ames2019control}. $\alpha \in (0,1)$ is the user-defined confidence level for Conditional Value at Risk calculation, e.g., $\alpha = 0.95$.
The $\mbox{CVaR}_{\alpha}(\cdot) \in \mathbb{R}$ calculation indicates the expected value of ($\cdot$) under the worst case of $5\%$ probability that something bad happens \cite{zhou2022risk}. The safety loss function $L_{ij}(x,{u})$ represents how close the system is to the boundary of the safe set given the user-specified confidence level $\alpha$, or how easily a safety violation could occur, under the assumption that both agents will move with piecewise-constant velocity.

Now with $L_{ij}(x,{u})$ as a handy tool describing the risk agent $i$ faces when interacting with agent $j$, for a multi-agent system, we define the aggregated risk $R_i \in \mathbb{R}$ agent $i$ faces posed by all surrounding agents as:
\begin{equation}
    R_i = \sum^N_{j=1} L_{ij}(x,{u}), \quad \forall j \neq i
    \label{risk-calculation}
\end{equation}
The larger $R_{i}$ is, the more likely safety violation is to occur. The proposed risk evaluation framework is simple yet effective: 1) $R_{i}$ grows with the increased number of agents in the system, as the environment becomes more complex and challenging; 2) $R_{i}$ varies depending on the changes of states, including positions and motion of other agents as we expected, as it is important to tell how much risk agent $i$ is exposed to even when collision has not happened yet.
\begin{figure}
    \centering
    \includegraphics[width = 0.8\linewidth]{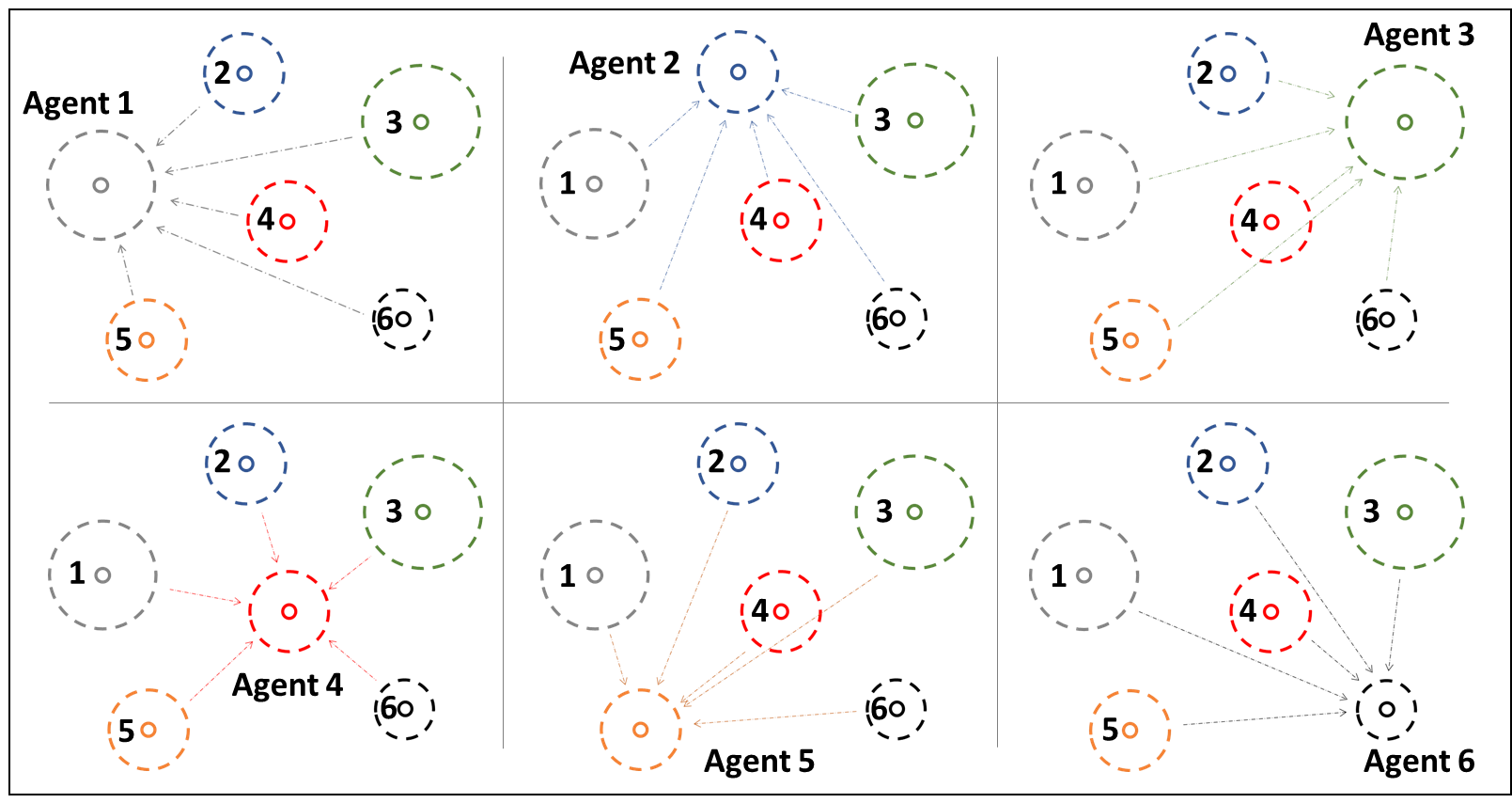}
    \caption{\footnotesize
    \label{mas-risk-measure}
   Propagated risk evaluation for individuals in a multi-agent system. Each arrow represents the risk posed by the pairwise relative movement under uncertainty of the evaluated agent and its neighboring agent.}
\end{figure}
Fig. \ref{mas-risk-measure} provides an illustrative example of how risk is calculated for individual agents in a multi-agent interaction scenario. For all six robots, the risk each individual agent faces consists of the pairwise risk generated by the surrounding five agents.

Note that the proposed risk evaluation framework does not necessarily require the agents to use Control Barrier Function-based controllers. We understand that in the real world agents may use different kinds of controllers, yet it does not prevent them from understanding the risk generated from multi-agent interaction via the proposed framework, with the mild but reasonable assumption that information about safety margin, agent states, and uncertainty distribution is known or observable. Even for agents not using CBF-based controllers, it is still possible to learn the parameter $\gamma$ from observations using machine learning techniques like linear ridge regression \cite{lyu2022adaptive}.

\subsection{CBF-inspired Risk Map for Multi-agent Interaction}
Based on the proposed risk evaluation framework, we now present CBF-inspired Risk Map (CBF-RM) as a visualization tool to better understand how risk is aggregated in multi-agent interactions.

\begin{figure}
    \centering
    \includegraphics[width = 0.8\linewidth]{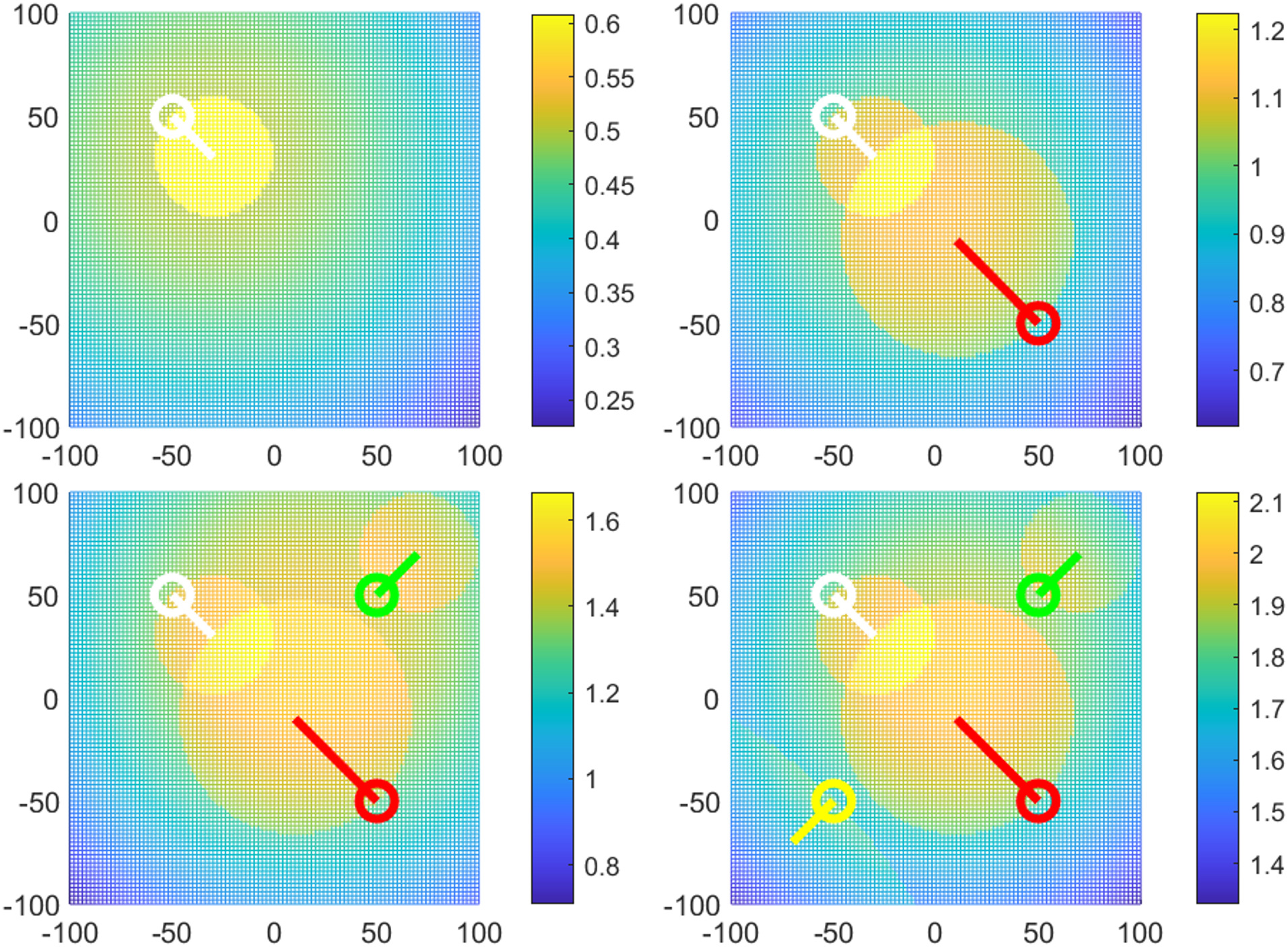}
    \caption{\footnotesize
    \label{risk-map}
   Control Barrier Function-inspired Risk Map (CBF-RM) showing how risk is aggregated when considering different numbers of agents in the system with various states and specifications. Agents are marked with different colors of circle and the lines pointing from circles represent agent velocity. The longer the line is, the larger velocity the agent is moving with.}
   \vspace{-0.5cm}
\end{figure}

To construct a map such as that shown in Fig. \ref{risk-map}, we augment the proposed agent-to-agent risk evaluation to agent-to-point risk evaluation, by assuming that for any point $p,\forall p \neq x_i \in \mathbb{R}^2, i\in \{i,...,N\}$ in the map, there exists a static agent with zero velocity. Then Eq. \ref{risk-calculation} can be used for augmented agent-to-point risk evaluation. Fig. \ref{risk-map} provides an example of how risk is aggregated when continuously adding agents with different positions, motion, safety radius, and parameter $\gamma$. Color ranging from dark blue to yellow represents the level of risk from mild to severe. Initially there is only one white agent and the highlighted yellow zone represents the collision zone, namely that if we place an agent with zero velocity anywhere inside this zone, collision will happen. As discussed earlier, we actually care more about the area outside the highlighted yellow zone, as that is the space within which robots with initial collision-free configurations navigate. It is observed that the closer the position to the agent, the greater the risk. When the red agent is added to the map, since it has a larger velocity, the risk it brings to the environment also increases. When adding the green agent to the map, driving away from the white and red agents, we observe that the risk in the area between three agents significantly increases. This makes sense as compared to the two-agent plot: if we put an agent at position (25,25), having the green agent on the side definitely brings more risk than without the green agent, even though it is driving away. In the last figure, a new agent is added with a smaller CBF parameter $\gamma$, indicating the agent behaves more conservatively with caution, and that's why the size of its collision zone is larger than other agents'. Considering white and red agents together, the risks posed by the green and yellow agents are different, even though they share the same speed, same distance away from white and red agents, and same safety radius. In this way, we show the proposed risk evaluation framework is generally applicable, as it embeds different factors including agent position, motion, safety radius, and behavior aggressiveness into one unified framework, thanks to the expressivity in the nature of CBF. 
In the following section, we choose CBF-based controllers as an example to demonstrate how to translate the centralized safe control into a decentralized setting, by dynamically allocating responsibility shares for pairwise agents based on CBF-inspired risk measurement.

\subsection{Decentralized Risk-aware CBF-based Controller}

We proved in our previous work~\cite{lyu2022responsibility} that we are able to compose formally provable decentralized CBF-based controllers by assigning responsibility shares $\omega$ to any pairwise agents $i$ and $j$, if we have known information about agent personalities $\phi$, where $\omega_i = cos^2(\phi_i)$. In doing so, we partition the admissible control space in the centralized constraint based on the agent personalities.

\textbf{Research Question 2}: However, without the known information of agent personalities, how can we assign such identities to agents, e.g., be a cautious agent or aggressive agent, in other words, how to allocate the responsibility shares $\omega$ between pairwise agents based on the proposed risk measurement under motion uncertainty?

For any pairwise agents $i$ and $j$, the centralized CBF-based safety constraint over agent velocity $u_i,u_j \in \mathcal{R}^2$ is in the linear form of:
\begin{equation}
    \mathcal{B}(x_i,x_j)=\{u_i\in\mathcal{U}_i,u_j\in\mathcal{U}_j: A_i(u_i- u_j) \leq b\}
    \label{centralized-cbf}
\end{equation}
where $A = -2(x_i-x_j)$ and $b = \gamma h(x_i,h_j) +2 \cdot \mbox{CVaR}_{\alpha}((x_i-x_j)^T(\epsilon_i-\epsilon_j))$.

\begin{theorem}
In a multi-agent system, agent safety during an interaction is formally guaranteed at a confidence level $\alpha$, if for any pair of agents $i$ and $j$, agent $i$ takes the \textbf{Pairwise Responsibility Weight} $\omega_i = \frac{R_j}{R_i+R_j}$, 
so that the admissible control space in a centralized system in Eq. \ref{centralized-cbf} is converted to Eq. \ref{decentralized-cbf}: 
\begin{equation}
    \begin{split}
    \mathcal{B}(x)=&\{u_i\in\mathcal{U}_i: A_i u_i\leq \omega_i b_i, 
    A_i = -2(x_i-x_j)^T \in \mathbb{R}^{1\times2}, \\b_i = \gamma& h(x_i,x_j) +2 \cdot \mbox{CVaR}_{\alpha}((x_i-x_j)^T(\epsilon_i-\epsilon_j)) \in \mathbb{R}\}    \end{split}\label{decentralized-cbf}.
\end{equation}
\label{respobsibility-theorem}
\end{theorem}
The main idea behind the weight design of $\omega_i$ is to compare the relative level of risk each set of pairwise agents is exposed to, so that the agent with lower risk can enjoy a wider admissible control space by taking a larger responsibility portion, compared to the agent with higher risk, which can be understood as its already being in a not-that-safe situation, and thus preferably only proceeding with caution with a tighter safety constraint. This design also reflects the idea of taking the neighbors of your neighbors into account, as for agent $i$'s weight calculation, $R_j$ embeds the information of constraints posed on agent $j$ by its own neighbors.

Therefore, for any pair of agents in multi-agent interaction, the \textbf{risk-aware CBF-based decentralized safe controller} is formulated as a quadratic program:
\begin{equation}
\begin{split}
      \min_{u_i\in\mathcal{U}_i}& || u_i-\bar{u}_i||^2 \\
    s.t \quad u_{min}& \leq u_i \leq u_{max}\\
    A_i u_i&\leq \omega_i b_i, \quad
    A_i = -2(x_i-x_j)^T,
    \omega_i = \frac{R_j}{R_i+R_j}\\    b_i = &\gamma h(x_i,x_j) + 2\cdot  \mbox{CVaR}_{\alpha}((x_i-x_j)^T(\epsilon_i-\epsilon_j))
\end{split}
\label{qp}
\end{equation}
where $\bar{u}_i \in \mathbb{R}^2$ is the nominal controller input, assumed to be computed by a higher-level task-related planner, for example, a behavior planner. $u_{min}$ and $u_{max}$ are the minimum and maximum allowed velocity. The objective function represents the goal of minimum deviation control, and the problem can be easily solved by a standard QP solver.

\begin{remark}(Solution feasibility)
In the presense of bounded input constraint, the above quadratic problem may become infeasible. This can be addressed by co-optimizing parameter $\gamma$ to enhance the non-emptiness of the solution set with bounded input constraint and CBF safety constraint \cite{lyu2021probabilistic}, or by enforcing an additional CBF constraint which characterizes the sufficient condition of solution feasibility \cite{xiao2022sufficient}.
\end{remark}

\begin{remark} (Long-term safety) Despite the fact that the proposed method is a step-wise optimization, the long-term probabilistic safety along an entire time horizon $[0,\tau]$ can be guaranteed if for any pairwise agents, $x_i,x_j\in \mathcal{H}_{ij}(x)$ and $u_i, u_j\in \mathcal{B}(x)$ for all $t\in [0,\tau]$.

\end{remark}

\subsection{Theoretical Analysis}

In this section, detailed proof is provided to show that the proposed framework in Th. \ref{respobsibility-theorem} is the sufficient condition of formally provable safety guarantees. In other words, we aim to show that for any pair of agents $i$ and $j$, Th. \ref{respobsibility-theorem} ensures the agents will not collide.

\begin{proof}

\textbf{Step 1:} By applying Th. \ref{respobsibility-theorem} to agent $j$, we get the safety constraint over $u_j$ as:
\begin{equation}
\begin{split}
    A_ju_j\leq\omega_j b_j, \quad A_j = -2(x_j-x_i)^T\\
    b_j = \gamma h(x_j,x_i) + 2\cdot  \mbox{CVaR}_{\alpha}((x_i-x_j)^T(\epsilon_i-\epsilon_j))
\end{split}
    \label{agentj-responsibility}
\end{equation}
We know $A_j = -2(x_j-x_i)^T = 2(x_i-x_j)^T$. The summation of the left hand sides of the inequalities in Eq. \ref{qp} and Eq. \ref{agentj-responsibility} is:
\begin{equation}
    \begin{split}
        A_i u_i + A_ju_j 
        = -2(x_i-x_j)^T u_i+2(x_i-x_j)^T u_j
    \end{split}
\end{equation}
which equals the left hand side of the inequality in Eq. \ref{centralized-cbf}.

\textbf{Step 2:} By Eq. \ref{safety-function}, we have $h(x_j,x_i) = h(x_i,x_j)$, $2\cdot \mbox{CVaR}_{\alpha}((x_i-x_j)^T(\epsilon_i-\epsilon_j))=2\cdot  \mbox{CVaR}_{\alpha}((x_j-x_i)^T(\epsilon_j-\epsilon_i))$, and therefore $b_i = b_j$. Now we sum the right hand sides of the two inequalities and get:
\begin{equation}\begin{split}
        &\omega_i b_i+ \omega_j b_j = (\omega_i+\omega_j) b_i\\
    &= (\frac{R_i}{R_i+R_j}+\frac{R_j}{R_i+R_j})*(\gamma h(x_i,x_j)\\
    &+2  \mbox{CVaR}_{\alpha}((x_i-x_j)^T(\epsilon_i-\epsilon_j))\\
    & = \gamma h(x_i,x_j)+2\cdot  \mbox{CVaR}_{\alpha}((x_i-x_j)^T(\epsilon_i-\epsilon_j))
\end{split}
\end{equation}
 which equals the right hand side of Eq. \ref{centralized-cbf}. Thus the proof is concluded that the proposed risk-aware decentralized CBF in Th. \ref{respobsibility-theorem} provides sufficient conditions to ensure formal safety guarantees.
\end{proof}

Next, the algorithm of the proposed framework is presented in Algorithm~\ref{alg:responsibility_ctrl}.
\begin{algorithm}\footnotesize
\caption{Risk-aware CBF-based \\Decentralized Safe Control Framework}
\begin{algorithmic}\label{alg:responsibility_ctrl}
\REQUIRE $x^0_{1,...,N}, \bar{u}^t_{1,...,N}, {u}^t_{1,...,N},\gamma, R_{safe},\epsilon_{1,...,N}$
\ENSURE $u^t_{1,...,N}$
\FOR{$t = 1:T$}
\FOR{$i=1:N$}
\FOR{$j=1:N$ except $i$}
\STATE
Perform risk evaluation for agent $i$ and $j$ (Eq. \ref{risk-calculation})
\STATE
Compute Pairwise Responsibility
Share $\omega_i$ (Th. \ref{respobsibility-theorem})
\STATE Calculate $A_i^t,b_i^t$ for decentralized safety constraint composition (Eq. \ref{decentralized-cbf})
\STATE Stack $A^t_i,b^t_i$ for all surrounding agent $j$
\ENDFOR
\STATE Minimum Deviation Control:  $\min_{u^t_i} || u^t_i-\bar{u}^t_i||^2$ \quad
\STATE
    s.t $\quad u^t_i \in [u_{min}, u_{max}],\quad A^t_i u^t_i\leq \omega_i b^t_i$
\ENDFOR
\ENDFOR
\end{algorithmic}
\end{algorithm}
The superscript $0$ represents the initial condition at $t = 0$, and the superscript $t$ represents variables at timestep $t$. At each timestep, we consider all pairwise agents for decentralized safety constraint composition. For every pairwise agent $i$ and $j$, the pairwise responsibility shares $\omega_i$ and $\omega_j$ are calculated based on the risks level two agents are exposed to. By allocating the responsibility shares dynamically, we allow the agent with lower risk to enjoy more freedom with a wider admissible control space, and constrain the motion of the agent at high risk with a tighter bound to force it to proceed conservatively with caution. The proposed framework scales up well with a larger number of agents and is highly generally applicable to other real-time robotics applications.

\section{Simulations \& Discussion}
\label{experiment}

In this section, we demonstrate the validity and effectiveness of our proposed risk-aware CBF-based decentralized controller from the following four aspects: 1) safety, 2) interpretability, 3) implicit coordination, and 4) task efficiency
through two examples.

\noindent\textbf{Example 1: Ramp Merging in Autonomous Driving}

The proposed risk-aware CBF-based decentralized controller is deployed on three autonomous vehicles in a ramp merging scenario where no over-the-air communication is available, shown on the left in Fig. \ref{responsibility-share}. With Vehicle 1 and 3 on the ramp and Vehicle 2 on the main road, the goal for each vehicle is to pass the merging point as quickly as possible while avoiding collisions with each other. For simplicity, we use V1, V2, and V3 to refer to the three vehicles.
The vehicle dynamics are described by double integrators as in \cite{lyu2021probabilistic}: $\dot{X} =\begin{bmatrix}
    \dot{x}\\
    \dot{v}
    \end{bmatrix}
    =\begin{bmatrix}
    0_{2\times2}\; I_{2\times2}\\
    0_{2\times 2} \;0_{2\times 2}
    \end{bmatrix}
    \begin{bmatrix}
    x \\
    v
    \end{bmatrix}
    + \begin{bmatrix}
    0_{2\times2}\; I_{2\times2}\\
    I_{2\times2}\; 0_{2\times2}
    \end{bmatrix}\begin{bmatrix}
    u\\ \epsilon
    \end{bmatrix}$, 
where $x\in\mathbb{R}^2,v\in \mathbb{R}^2$ are the position and velocity of each vehicle. $u\in\mathbb{R}^2$ represents the acceleration control input that is being optimized (Algorithm~\ref{alg:responsibility_ctrl}). $\epsilon \in \mathbb{R}^2$ is a random Gaussian variable with known distribution, representing the motion uncertainty in agent velocity.

\subsection{Safety Validation}
\begin{figure}
    \centering
    \includegraphics[width = 1\linewidth]{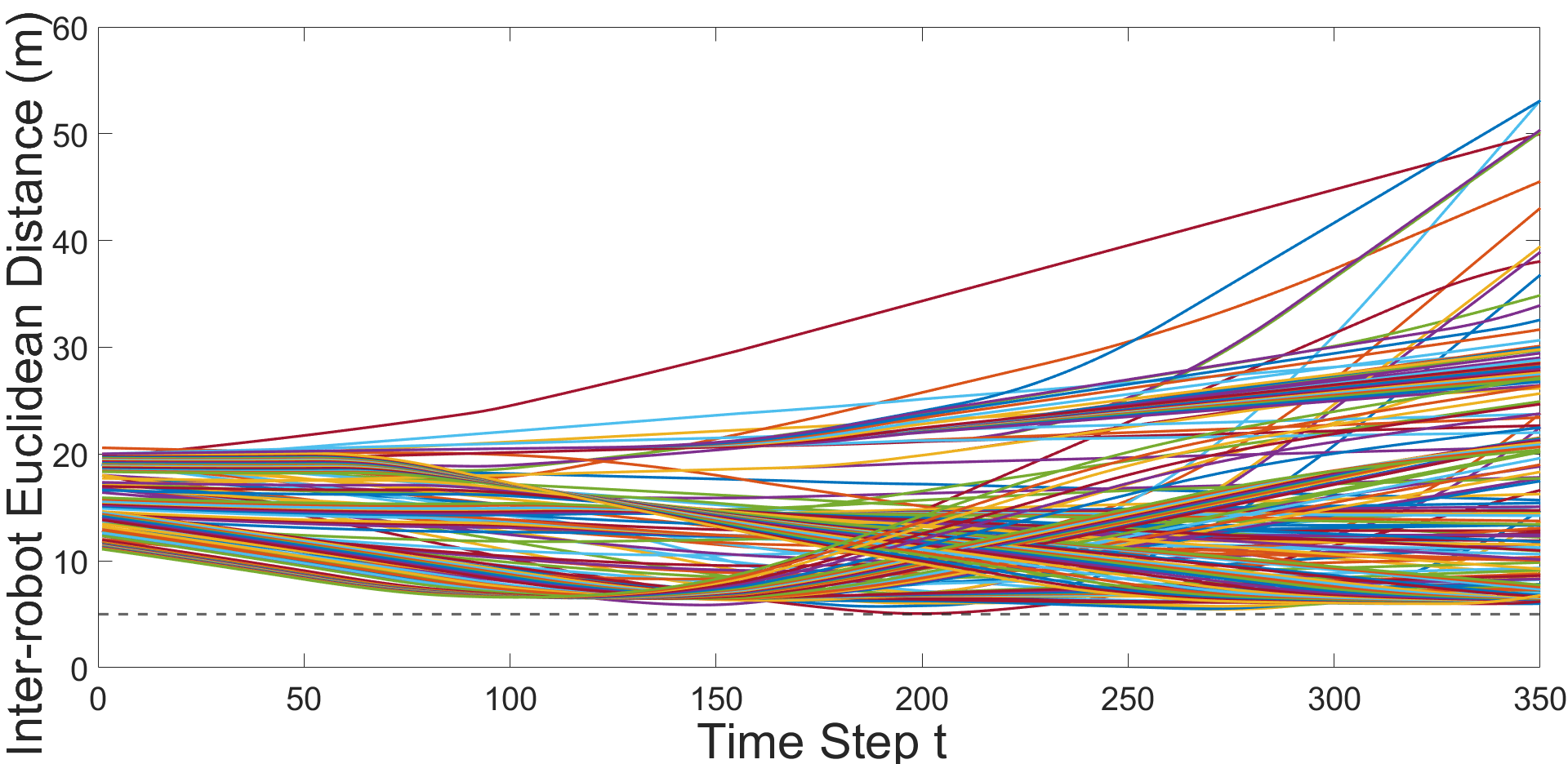}
    \caption{\footnotesize
    \label{20trials}
   Euclidean distance between three pariwise vehicles in 50 trials over time with randomized initial configurations in the ramp merging scenario.}
   \vspace{-0.5cm}
\end{figure}

To validate the safety performance of our proposed method, we set the confidence level $\eta= 99.9\%$. 50 trials are conducted with randomized initial configurations, including vehicle position, velocity and acceleration.  The Euclidean distance between all the $50\times3 = 150$ pairwise relationships over time is recorded in Fig. \ref{20trials}.  The grey dashed line represents the safety margin $R_{safe}=5$ m. 
It is observed that the inter-vehicle distance in all trials is staying above the safety margin, meaning that the safety requirement is being satisfied and no collision happens.

\subsection{Physical Interpretation}
One of the advantages of the proposed method is its yielding a measure of relative safety and adaptive conservativeness that is interpretable. The right plot in Fig. \ref{responsibility-share} provides an example in which V2 accelerates to merge into the gap between V1 and V3. The figure depicts how the three vehicles adapt their behavior conservativeness based on the relative difference of risk they receive in the interaction.

The situations in the top subplot and the bottom one are similar. Taking the top subplot as an example, when analyzing the responsibility shares between V1 and V2, the presence of V3 brings an increasingly higher risk to V2 while it's merging into the gap compared to V1. Therefore, V2 is assigned a continuously decaying responsibility share compared to V1, meaning its motion is constrained more tightly. This aligns with our expectation that since V2 is on the main road at first with no other vehicles in the front or back, it enjoys a lot of freedom. However, as V2 completes the merge, it has to be very careful in the action it chooses to avoid colliding with V1 or V3.


In the middle subplot, we observe that the responsibility shares between V1 and V3 experience two swaps. The first swap happens when V2 is about to merge into the gap. V2's merge brings V3 higher risk compared to V1, as the risk V2 generated is highest along the direction it moves. Therefore, it's not surprising that the two vehicles' responsibility shares swap around $t=200$. The increasing pressure on V3 posed by V2 makes V3 keep accelerating as its reaction, trying to maintain a further distance from V2. Then with the increased distance between V2 and V3, the relative risk V3 receives decreases significantly compared to V1, causing the second swap.

\noindent\textbf{Example 2: Multi-agent Position Swapping Task}

In this example, we have six agents in total with the task of swapping positions with each other. All agents aim to safely navigate to their goal locations, employing a move-to-goal controller $\bar{u} = -k\cdot(x-x_{target})$, where $k\in \mathbb{R}^{>0}$, and $x_{target}\in\mathbb{R}^2$ is the goal position of each agent. A right-hand heuristic rule is used for deadlock resolution, as in \cite{pierson2020weighted}. We compare our proposed method with 1) a centralized control baseline approach and 2) a non-risk-aware decentralized baseline approach.

\subsection{Comparison with Centralized Control}
In this section, we compare our proposed decentralized CBF-based control approach with the centralized CBF-based control approach. In the centralized CBF-based control approach, the problem is formulated as a single optimization, trying to calculate the optimal control for all six agents at the same time. The objective is to minimize the overall control deviation from the nominal controller of all agents while respecting all the pairwise safety constraints in Eq. \ref{centralized-cbf}. 

\begin{figure}
    \centering
    \includegraphics[width = 1\linewidth]{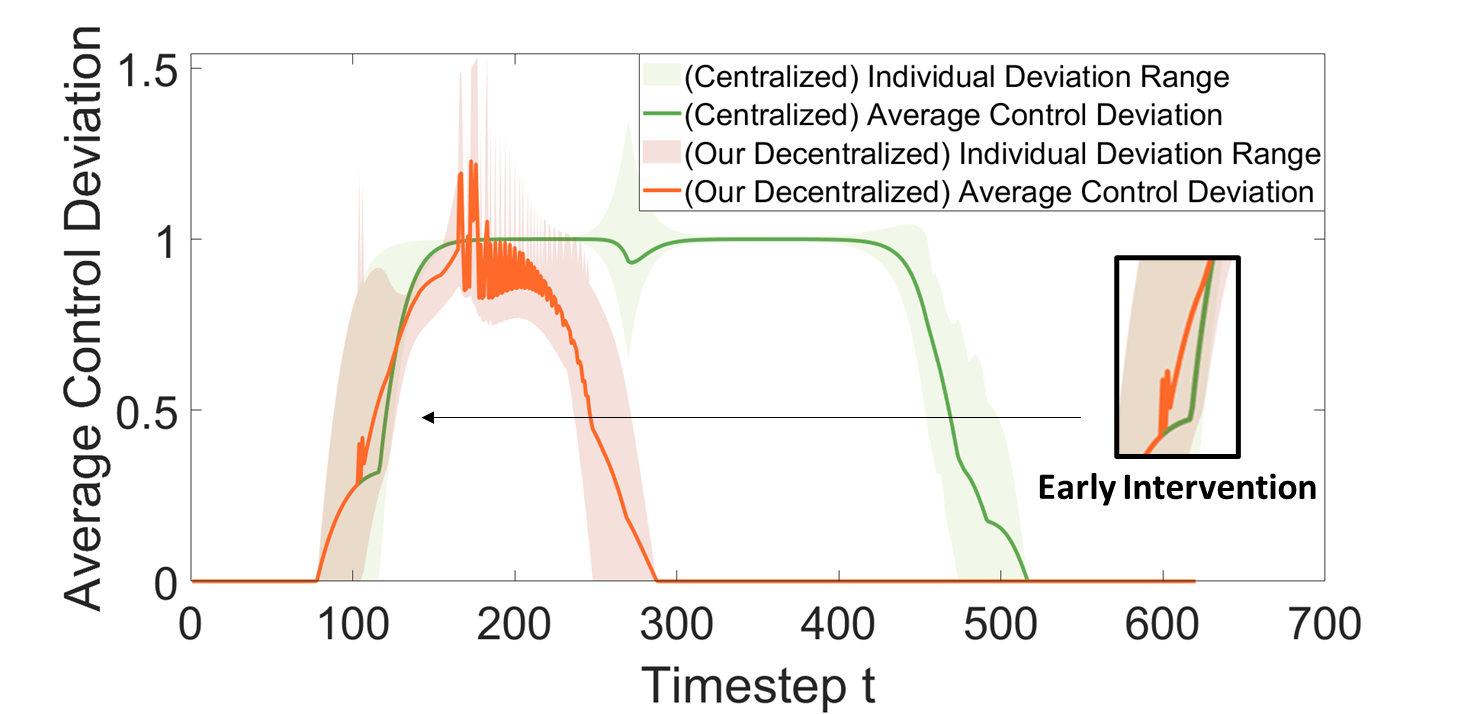}
    \caption{\footnotesize
    \label{centralized-decentralized-comparison}
   Robot control deviation comparison of our proposed decentralized control with centralized control.}
   \vspace{-0.5cm}
\end{figure}

The average control deviation among the agents during the whole task is depicted in Fig. \ref{centralized-decentralized-comparison}. The orange line and the green line represent the average control deviation from the nominal controller using our proposed decentralized CBF-based control and the centralized CBF-based control respectively. The shaded regions represent the maximum and minimum individual deviation over time. It is observed that although the maximum deviation of our decentralized method is greater than that of the centralized method, the overall duration of control deviation is about 44\% shorter than the centralized one. This means with our method, less intervention is applied to the agents to complete the task safely. Our proposed method also completed the task faster than the centralized method. Their time of task completion is 402s and 617s respectively.

Through this comparison, we can see that in the absence of an explicit global coordinator, the dynamic responsibility allocation process in the proposed method plays the role of an implicit coordinator among agents. This makes early intervention possible (enlarged in the figure), forcing robots with higher risk to take precautionary measures even when agents are not close enough to collide immediately. This results in heterogeneous behavior among agents that helps to avoid possible deadlock situations and complete the task faster.

\subsection{Comparison with Non-risk-aware Decentralized Control}
We compare our proposed risk-aware decentralized CBF-based control approach with the non-risk-aware decentralized CBF-approach in terms of average control deviation from the nominal controller, the task efficiency as well as the agent behavior observations. The non-risk-aware approach is implemented with the same CBF-based decentralized controller, but with fixed and equal responsibility shares among agents, representing their non-changeable behavior conservativeness regardless of the amount of risk agents receive.

\begin{figure}
    \centering
    \includegraphics[width = 1\linewidth]{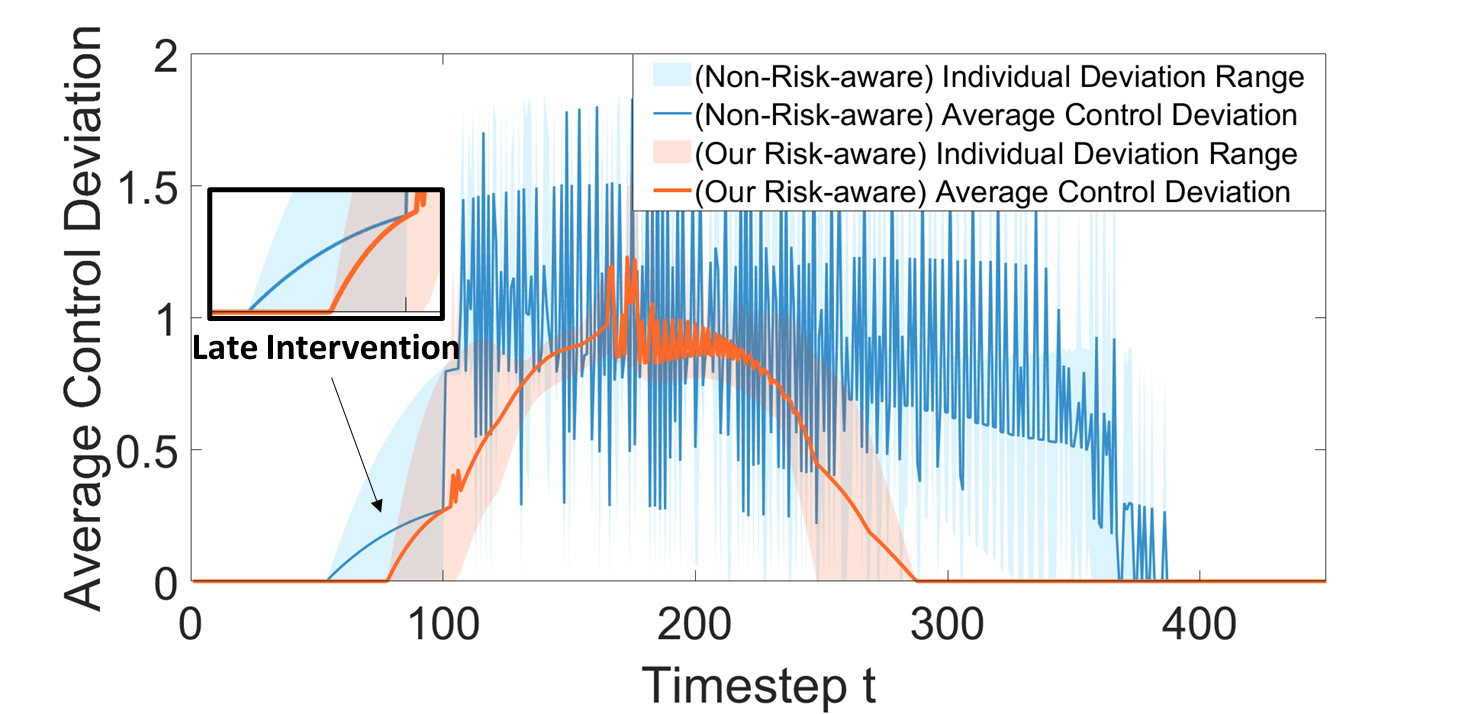}
    \caption{\footnotesize
    \label{risk-nonrisk-comparison}
   Robot control deviation comparison of our proposed risk-aware decentralized control and non-risk-aware decentralized control in robot control deviation.}
   \vspace{-0.5cm}
\end{figure}

The average control deviation during the whole task is recorded in Fig. \ref{risk-nonrisk-comparison}. It is observed that our proposed method has a lower maximum individual control deviation and a shorter overall duration of intervention compared to the non-risk-aware method. In the non-risk-aware method, all agents are assigned fixed and equal responsibility shares, meaning they all have the same tight control constraint, no matter whether the agent is in a risky or extremely safe situation. This leads to unnecessary conservative behavior, for instance, agents have to decelerate significantly even when they are very far away from each other. Thanks to the risk-aware responsibility allocation in our method, late intervention (enlarged in the figure) ensures that nominal controllers are only revised when necessary. Collective safety is achieved at a lower cost of sacrificing control optimality with our method.

\begin{figure}
 \center
  \includegraphics[width=1\linewidth]{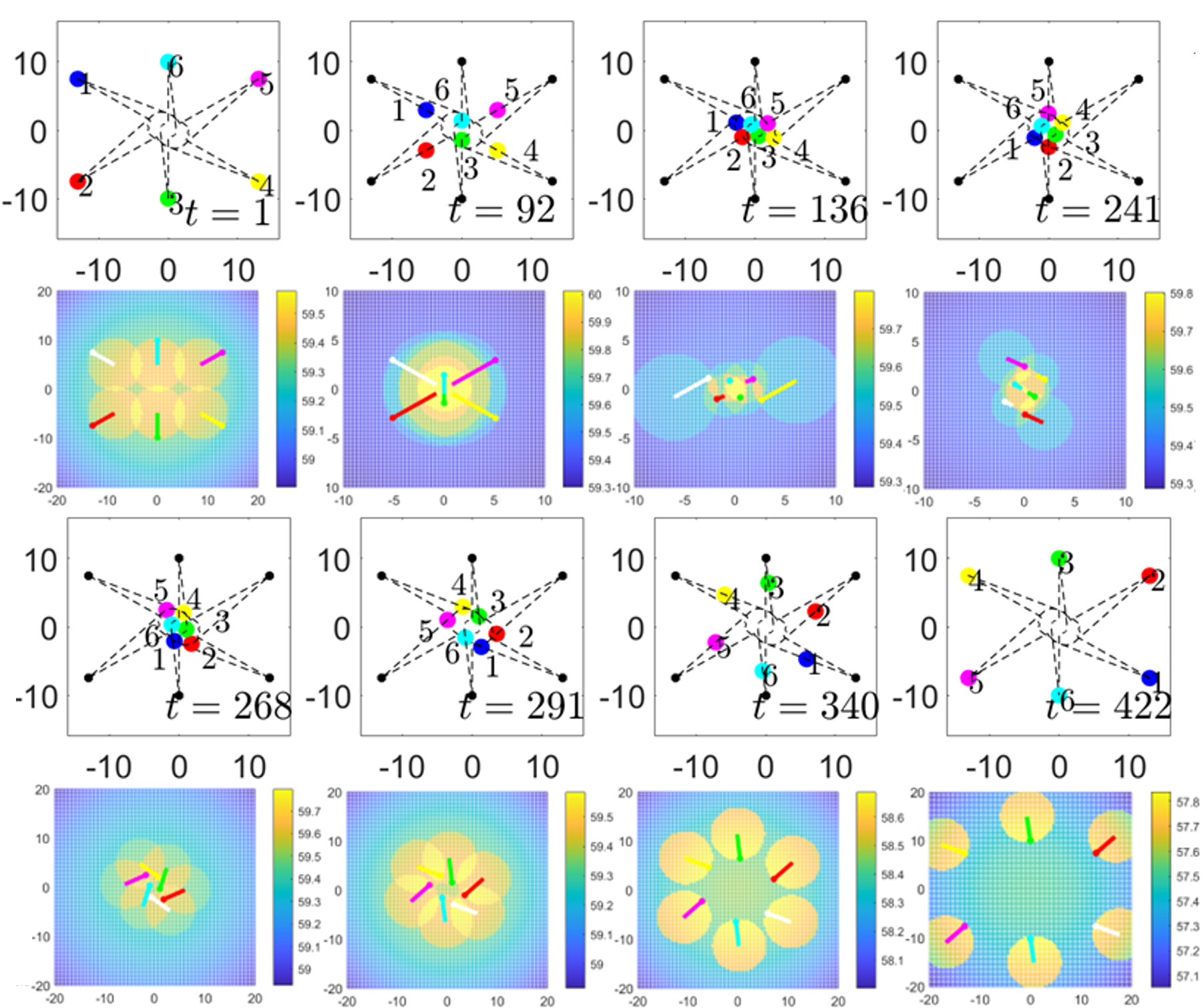}
  \caption{Multi-agent position swapping game with fixed and equal responsibility allocation (Baseline method).}
  \label{mas-swap-equal}
\end{figure}

\begin{figure}
 \center
  \includegraphics[width=1\linewidth]{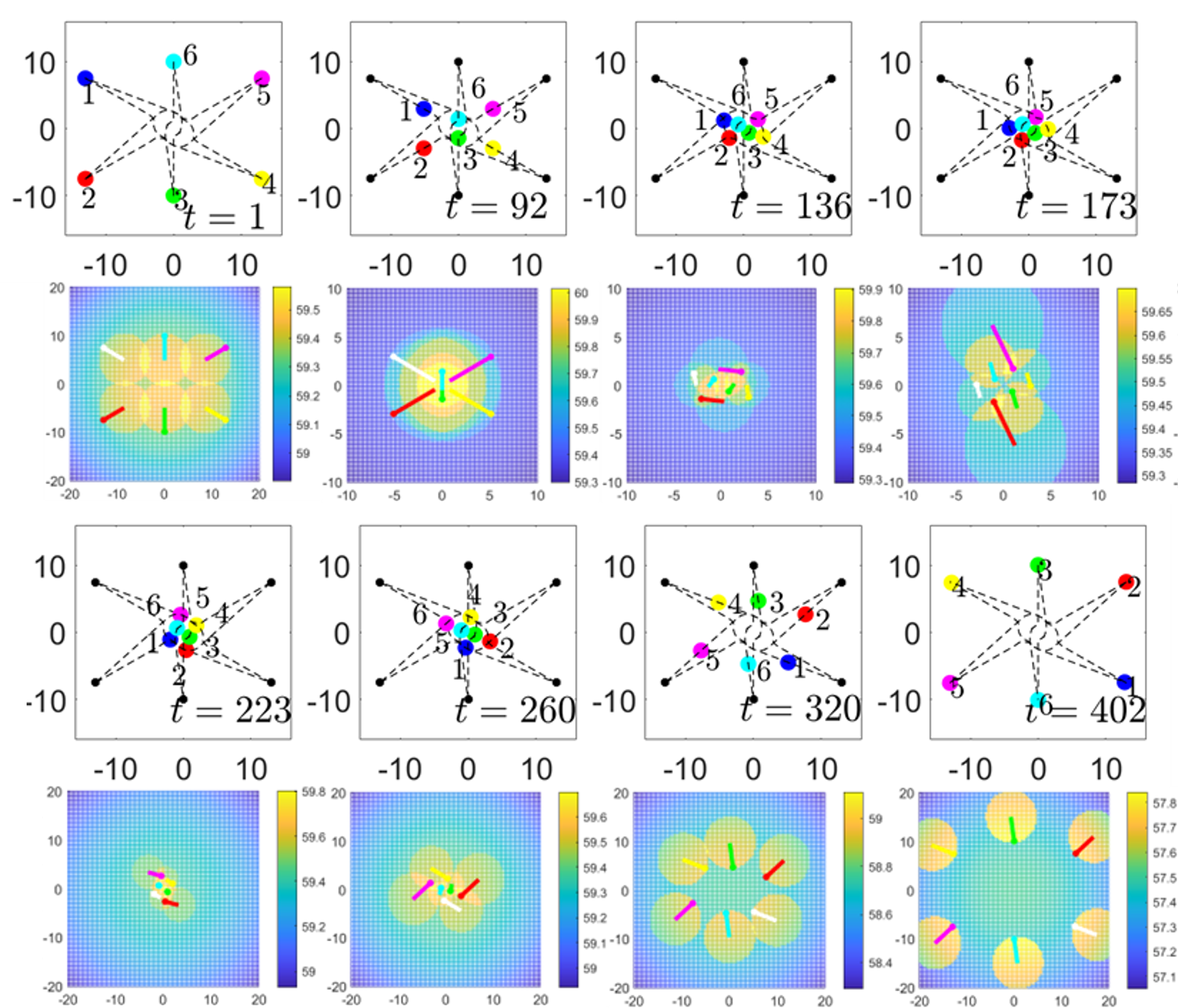}
  \caption{Multi-agent position swapping game with dynamic responsibility allocation (proposed method).}
  \vspace{-0.5cm}
  \label{mas-swap-dynamic}
\end{figure}

Next, comparisons of task efficiency and agent behavior observations are shown in Fig. \ref{mas-swap-equal} and Fig. \ref{mas-swap-dynamic}. It is shown that our proposed method improves the overall task efficiency by 4.7\% with less time spent. 

To better analyze the difference in agent behavior in the two methods, we use the proposed CBF-RM for visualization. 
 Since the initial and final agent configurations are the same, the first two and last two subplots in the two cases are almost the same. Agents come closest to one another in both cases starting in the third subplot, and start to form a rotational formation due to the right-hand heuristic for deadlock resolution. In the baseline method, from $t=241$ to $t=268$, since all agents are assigned equal responsibility shares, agents 3 and 6 surrounded in the middle are not additionally constrained compared to others and therefore try to break the formation and escape from being surrounded in the middle. On the other hand, at the same time in our proposed approach, since agents $3$ and $6$ are recognized as the agents experiencing the highest risk level, they are automatically assigned very small responsibility shares, resulting in a very limited admissible control space. Therefore, they have no choice but to stop where they are and wait for other agents to complete the formation rotation. Surprisingly, having the riskiest agents stop and wait in our method actually brings higher overall task efficiency, as tending to escape from where they are in the baseline method just makes things worse and prevents the agents with higher flexibility in motion from completing their job smoothly. Therefore from this point of view, our proposed method is a promising way to configure heterogeneous robot teams, as the way it assigns responsibility shares implicitly encourages agents to be more cooperative instead of competitive when facing conflict.

\section{CONCLUSIONS}
\label{conclusion}
We proposed a CBF-inspired risk evaluation framework to measure the aggregated risk individual agents face due to multi-agent interactions under motion uncertainty. By leveraging this measurement, responsibility shares are dynamically allocated among agents to construct CBF-based decentralized controllers with adaptive motion flexibility and behavior conservativeness. The resulting risk-aware decentralized CBF-based safe controller is shown to be valid and effective in simulations. The augmented CBF-RM works as a helpful visualization tool to help explain the aggregation of risk in complex dynamic environments and could be potentially useful in CBF-RM-aided controller design in the future.

\bibliography{ICRA}{}
\bibliographystyle{IEEEtran}

\end{document}